\pdfoutput=1

\documentclass[11pt]{article}

\usepackage[]{ACL2023}

\usepackage{times}
\usepackage{latexsym}

\usepackage[T1]{fontenc}

\usepackage[utf8]{inputenc}

\usepackage{microtype}
\usepackage{adjustbox}
\usepackage{inconsolata}
\usepackage{amssymb}
\usepackage{graphicx}
\usepackage{booktabs}

\title{Evaluating Pragmatic Abilities of Image Captioners on A3DS}

\author{Polina Tsvilodub \and Michael Franke \\
  Department of Linguistics \\
  University of Tübingen \\
\texttt{\{polina.tsvilodub, michael.franke\}@uni-tuebingen.de}
  }

\begin{document}
\maketitle
\begin{abstract}
Evaluating grounded neural language model performance with respect to pragmatic qualities like the trade off between truthfulness, contrastivity and overinformativity of generated utterances remains a challenge in absence of data collected from humans.
To enable such evaluation, we present a novel open source image-text dataset ``Annotated 3D Shapes'' (A3DS) comprising over nine million \textit{exhaustive} natural language annotations and over 12 million variable-granularity captions for the 480,000 images provided by \citet{burgess20183d}. 
We showcase the evaluation of pragmatic abilities developed by a task-neutral image captioner fine-tuned in a multi-agent communication setting to produce \textit{contrastive} captions. The evaluation is enabled by the dataset because the exhaustive annotations allow to quantify the presence of contrastive features in the model's generations. We show that the model develops human-like patterns (informativity, brevity, over-informativity for specific features (e.g., shape, color biases)). 
\end{abstract}

\section{Introduction and Related Work}
\label{section1}
In human communication, language is rarely used as a unimodal channel; rather, language is mostly used in reference to the surroundings, i.e., it is \textit{grounded} in the physical world. Thus, in order to build artificial agents that could be potentially employed in scenarios requiring natural communication with humans, it is crucial to develop approaches for training such agents to communicate about the world in a human-like way \citep{lake_ullman_tenenbaum_gershman_2017}. However, automatically evaluating the human-likeness of a trained system without costly human feedback is a recurring problem in NLP. 

In this paper, we set out to provide tools for evaluating human-like pragmatic abilities of grounded models and evaluate a model trained interactively via reinforcement learning, which is commonly suggested to give rise to task-oriented behavior \citep{Lazaridou2020EmergentMC}.

Grounding of neural language models has been advanced greatly in recent years through \textit{image captioning models}. Starting with the work by \citet{vinyals2016show} and \citet{karpathy2014deep}, neural encoder-decoder architectures have been dominating the field, recently extending to unified architectures \citep{zhou2020unified}.
However, these approaches are \textit{task neutral}, i.e., the models are trained to produce generally true image captions. 

In contrast, humans are highly flexible and \textit{pragmatic} in their use of language and, e.g., adapt the granularity of their utterances to the requirements of the communicative task \citep{searle1969speech}. It is generally guided by conversational maxims, suggesting that cooperative speakers should only provide as much information as required in a given context, be truthful, relevant, and brief \citep{grice1975logic}. Therefore, faced with a simple referential task of picking out a target item among an array of distractors, humans tend to mention \textit{contrastive} features of the target \citep[e.g.,][]{KramerDeemter2012:Computational-G}, i.e., the ones setting it apart from distractors. On the other hand, \textit{biases} towards producing shape and color descriptions even when these aren't contrastive have been identified \citep[e.g.,][]{degen2020redundancy}.
For grounded language models, the underlying pragmatic reasoning formalized as nested Bayesian inference about the behavior of speakers and listeners \citep{goodman2016pragmatic} inspired decoding schemes applied on top of standardly trained models \citep[e.g.,][]{cohn-gordon-etal-2018-pragmatically, zarriess-etal-2021-decoding, shen-etal-2019-pragmatically, vedantam2017context, andreas-klein-2016-reasoning}. 

However, evaluating the pragmatic qualities of models' predictions when they are applied to specific tasks (e.g., referential tasks) remains a challenge.
Currently standard metrics like BLEU-n, ROUGE, CIDEr and METEOR \citep{papineni-etal-2002-bleu, banerjee-lavie-2005-meteor, vedantam2015cider, lin-2004-rouge} for evaluating models' generations make reference to the surface form of ground truth image annotations. They cannot provide insight into models' mechanics and possible biases based on \textit{context-dependent functional aspects} like mentioning contrastive features or being overinformative. 
Given that model predictions might not always be syntactically well-formed and yet still count as functionally expedient for a human (e.g., see Fig.~\ref{fig:example}), evaluating pragmatic aspects of natural language image captions is important. We propose a new dataset and metrics facilitating such evaluation in the next sections. 

\section{Methods}
\subsection{A3DS}
\begin{figure}[t]
\centering
    \includegraphics[scale = 0.2]{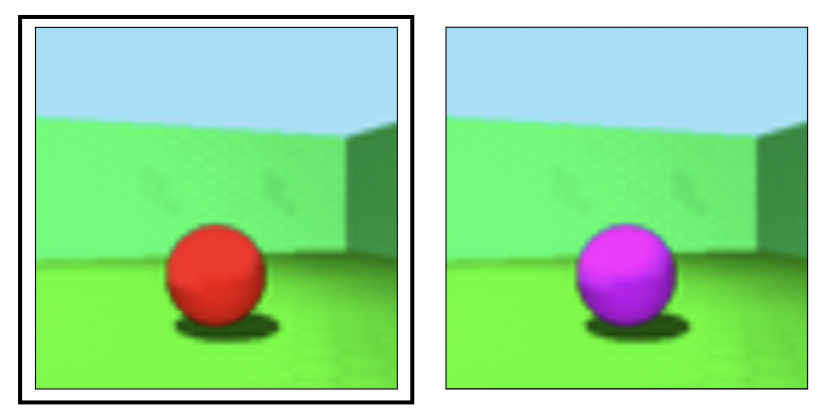}\\
    \caption{Example image pair matching on five features (left: red, right: purple ball), left image is target. Example exhaustive ground truth caption for target: ``A tiny red ball near the right corner in front of a light green wall on green floor.'' Example short ground truth caption: ``A ball on green floor.'' Contrastive caption predicted by RP model: ``A tiny \textbf{red} ball green near the floor in green of''.\footnotemark }
    \label{fig:example}
\end{figure}
\footnotetext{The last token was predicted nine times. This shows how the caption can be contrastive for the given task inspite of surface form artefacts.}

To enable such evaluation, we provide novel annotations for the dataset 3DShapes \citep{burgess20183d} (introduced in \citet{pmlr-v80-kim18b}) in the ``Annotated 3D Shapes'' (A3DS) dataset. The image dataset consists of 480,000 unique images of 3D geometric objects, constructed by varying six features ($\times$number of distinct feature values): shape type ($\times$4), shape color ($\times$10), shape scale ($\times$8), shape orientation relative to the background ($\times$15), wall color ($\times$10) and floor color ($\times$10). For each image, two sets of ground truth captions were generated: \textit{exhaustive} captions mentioning all six features and their values, and \textit{short} captions, mentioning two or three features of the image only (see example annotation in Fig.~\ref{fig:example}). The captions were constructed with a hand-written grammar from the numeric labels shipped with the original dataset. For each distinct feature value, different natural language descriptions were created. In total, over nine million exhaustive captions and 12 million short captions are released as part of this work.\footnote{\url{https://tinyurl.com/2p8w6rct}. The repository also contains endpoints for running model evaluations described in the next section and a sandboxed version of the dataset and the pretrained model for easy exploration.} 
The important advantage of this synthetic dataset for investigating referential language use of models trained on it is that the numeric labels allow to easily identify \textit{contrastive} versus \textit{redundant} features of the target image in any given context of distractor images. Furthermore, training with fully exhaustive captions allows to focus evaluations on models' contrastive abilities, excluding insufficient granularity of training data as a potential reason for a system's failure to be contrastive. 

Because all natural language expressions for each label are known, it is possible to comprehensively evaluate model predictions by-feature.
Predictions of fine-tuned models which may deviate from ground truth captions in their surface form (e.g., due to language drift; see, e.g., \citet{lazaridou-etal-2020-multi}) can also be evaluated.
We consider a caption contrastive if at least one of the known contrastive features for a given context (target and distractors) is mentioned in the target's description.
For contrastive color features, a caption is considered contrastive if it mentions the respective color irrespective of other mentioned aspects, if the color is unique for the target.
If several features in the target image have the same color, the description is considered contrastive only if the color name occurs together with the correct head noun (e.g., "floor", "wall", object shape).
For other contrastive features like shape, the respective expression (e.g., "ball", "in the left corner") has to literally occur in the generated caption.
For the example, in Fig.~\ref{fig:example}, we were able to identify that the caption is contrastive because the contrastive feature is the red color of the ball in the target image (left), there is only one red feature in the target image, and the generated caption contains the term "red".

We suggest informative metrics for evaluating pragmatic abilities of models on this dataset in the next section.

\subsection{Evaluation Metrics}
The metrics are informed by notions that are considered important in the cognitive science literature for cooperative and efficient pragmatic communication \citep[e.g.,][]{grice1975logic} and used commonly in the literature on computational generation of referring expressions \citep[e.g.,][]{KramerDeemter2012:Computational-G}. 
In the context of a reference task, we define pragmatically relevant categories of features a model might mention. Given a target and distractor image, each feature falls in one of the following three categories:
\begin{itemize}
\item \textit{Contrastive} feature: true of target and false of distractor.
\item \textit{Non-contrastive} feature: true of both the target and the distractor, and, therefore, redundant for the purpose of reference. 
\item \textit{False} feature: false of the target.
\end{itemize}
From these categories, we derive the following metrics (higher values are better), where $c$ is the number of contrastive features mentioned in a generated caption $y$, $k$ is the total number of features mentioned in $y$, and $z$ is the ground truth number of contrastive features between the images:
\begin{itemize}
    \item \textit{Discriminativity} $d$: $d = 1$ if $c > 0$ else 0, indicating if the caption successfully identifies the target,  thus a binary measure of task success.
    \item \textit{Contrastive efficiency} $e$ (applies only to  discriminative captions, i.e., for $d = 1$): $e = 1$ if $k = c = 1$, else: $e = 1 - \frac{c-1}{k-1}$, indicating whether the description avoids overmodification with contrastive features. This notion captures the extent to which the caption is economic and observes the communicative Maxim of Quantity, i.e., includes necessary details for the task but not more \citep{grice1975logic}.
    \item \textit{Relevance} $r$: $r = 1 - \frac{k - c}{6-z}$, indicates the propensity to avoid producing redundant non-contrastive features. This is formalized via the proportion of mentioned non-contrastive features ($k - c$) compared to all non-contrastive features ($6 - z$). It represents the communicative Maxim of Relevance \citep{grice1975logic} by measuring the degree to which details unnecessary for the task are excluded.
    \item \textit{Optimal discriminativity} $od$: $od = 1$ if $c=1$ else 0. It is a binary indicator summarizing $d$ and $e$, by binarizing the observance of the Maxim of Quantity for contrastive captions only \citep{grice1975logic}.
\end{itemize}
In the next section, we showcase how these metrics can be applied in order to evaluate the development of pragmatic abilities of an image captioner through fine-tuning in an interactive setting.

\begin{table*}[ht]
\centering
\begin{tabular}{llllllllll}
\toprule
& \multicolumn{3}{c}{one feature}  
& \multicolumn{3}{c}{two features}  
& \multicolumn{3}{c}{three features} \\ \cmidrule(r){5-7} \cmidrule(r){2-4} \cmidrule(r){8-10}
\textbf{Score} & \textbf{B} & \textbf{RP} & \textbf{SP} & \textbf{B} & \textbf{RP} & \textbf{SP} & \textbf{B} & \textbf{RP} & \textbf{SP}\\
\midrule
Discriminativity & \textbf{0.999} & 0.822 & 0.824 & 0.997 & 0.576 & 0.586 & 0.984 & 0.527 & 0.541\\
Contrastive efficiency & 0.198 & \textbf{0.879} & 0.875 & 0.203 & 0.963 & 0.955 & 0.251 & 0.856 & 0.875 \\
Relevance & 0.150 & 0.668 & 0.640 & 0.162 & 0.522 & 0.521 & 0.149 & \textbf{0.684} & 0.665\\
Optimal contrastivity & 0.014 & \textbf{0.457} & 0.452 & 0.039 & \textbf{0.485} & 0.476 & 0.148 & 0.335 & 0.367\\
Mentioned features \# & 5.880 & 2.944  & 3.125 & 5.871 & 2.950 & 3.133 & 5.876 & 2.955 & 3.135 \\
Listener accuracy & --- & 0.919 & 0.895 & --- & 0.887 & 0.900 & --- & 0.862 & 0.860 \\
\bottomrule
\end{tabular}
\caption{Pragmatic evaluation results by test set category for each model (B: pretrained baseline, RP: random pairs fine-tuning, SP: similar pairs fine-tuning), averaged across test sets within category. Bold numbers indicate best performance across models and test sets.}
\label{tab:pred-eval}
\end{table*}

\begin{figure*}[ht]
\centering
    \includegraphics[scale = 0.25]{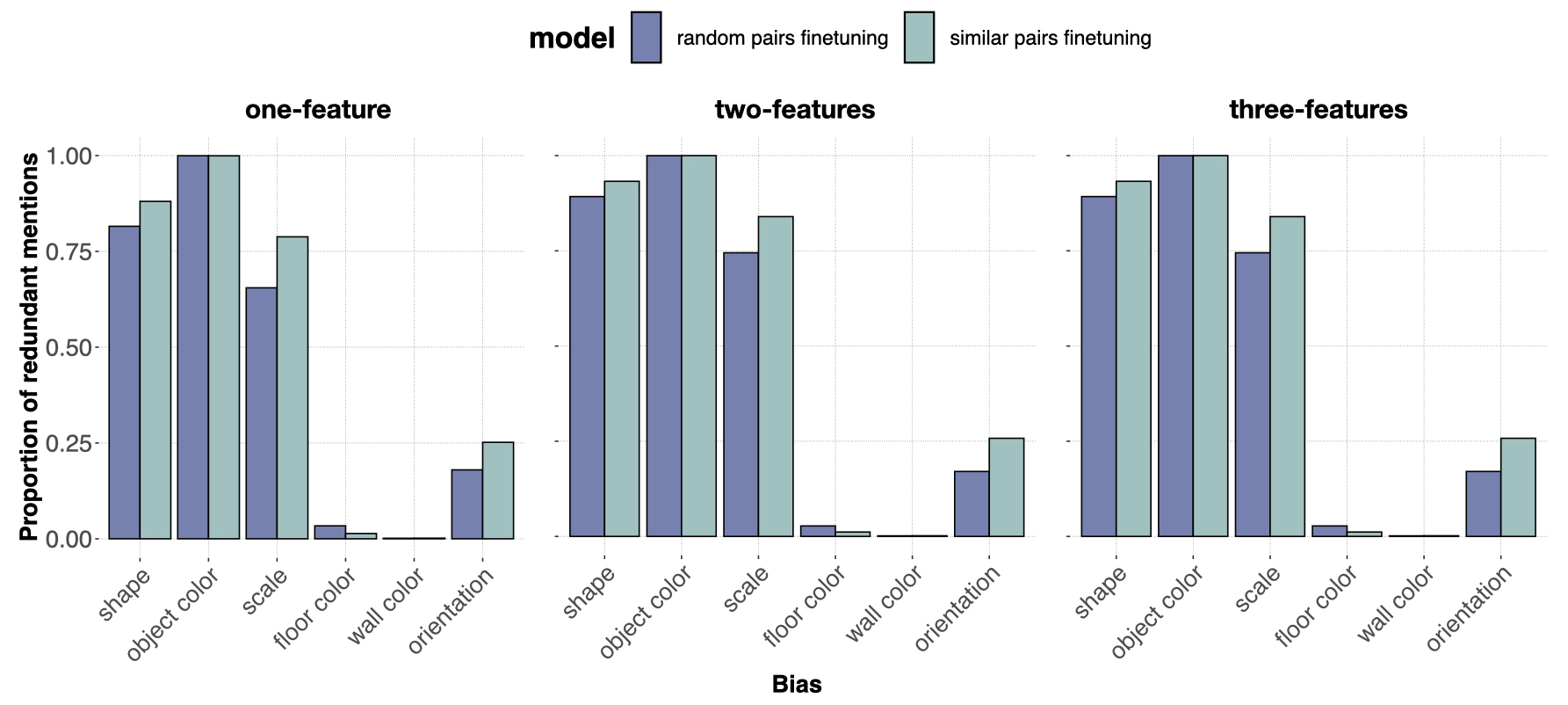}\\
    \caption{Generation proportions of each feature (x-axis) when it was \textit{non-contrastive} for each model (color) by test category (facets). Generation proportions of all features for the baseline (not shown) are at ceiling on all test sets, except for the scale category being at around 0.9 due to a tokenization glitch. \label{fig:byFt-biases}}
\end{figure*}
\subsection{Experiment}
The multi-agent communication setting wherein the image captioner is trained as the sender agent together with an artificial receiver agent to complete a communicative task (e.g., reference game) allows to fine-tune the sender's captioning behavior based directly on task performance, e.g., via \textit{deep reinforcement learning} \citep[e.g.,][]{lazaridou-etal-2020-multi, Lazaridou2020EmergentMC, lazaridou2016multi, havrylov2017emergence}, without making use of a supervised task specific dataset. Applied to the reference task, the idea is that the sender agent will learn to produce more contrastive descriptions which are helpful for the receiver to complete the task. \citet{lazaridou-etal-2020-multi} compare sender agent architectures in terms of their task-specific improvement, 
but they do not investigate properties like overinformativity that might have emerged during the multi-agent training. 

To investigate these potentenial effects, following the ``multi-task learning'' training regime from \citet{lazaridou-etal-2020-multi} we pretrained a \textit{baseline} image captioner (B) on 150,000 image-exhaustive caption pairs constructed from 30,000 images sampled from A3DS. It was then fine-tuned on another 150,0000 pairs on a reference game together with a listener agent. In the reference game, both agents received concatenated pairs of images $i = [i_1; i_2]$, where $i_t, t \in \{1,2\}$ was the target known only to the sender. The sender was trained to produce a description of the target, so that the listener guesses the target correctly, given the same images in randomized order. The sender received the reward $r=1$ if the guess was correct, and $r=-1$ otherwise. Both the sender and the listener consisted of a pretrained ResNet-50 image encoder which was not fine-tuned during the reference game, and a trainable linear layer projecting the ResNet image features to 512-dimensional features. These were input into one-layer LSTM language modules with the hidden layer size $h=512$. Further architectural and training details followed \citet{lazaridou-etal-2020-multi}.\footnote{The weight $\lambda_s$ for the speaker loss was set to 0.75.} 

We trained two sender-agent pairs in the reference game setting: in the \textit{random pairs} setting (RP), the agents saw pairs of (distinct) images selected at random. In the \textit{similar} pairs setting (SP), they received images which had at least three overlapping features (e.g., target and distractor depicted the same shape of the same color with background of the same color).\footnote{The speaker included $\approx$5.3M, the listener $\approx$2.15M trainable parameters. Pretraining for 10 epochs took around 10h, fine-tuning for 5 epochs---20h/model on NVIDIA A40 GPU.}

\section{Results}
The agents were evaluated on three categories of test sets, each set containing 7500 image pairs. In the \textit{one-feature} category, six sets were constructed where test pairs matched at least on one of each possible features. The \textit{two-features} category included three sets of pairs matched on at least two object features and a set with two random matching features. The \textit{three-features} category included sets where at least all object features, all background features, or three randomly sampled features matched. These sets allowed to evaluate in which conditions it was more difficult for the sender to produce appropriate captions.
In the following, the fine-tuned sender models (\textbf{RP} and \textbf{SP}) are compared to the baseline model (\textbf{B}), which is the pretrained task-neutral image captioner. The average number of falsely named features was 0.720 for baseline, 0.139 (RP) and 0.316 (SP). Table \ref{tab:pred-eval} shows listener test accuracies on all test splits, showing that the agents successfully learned the reference task (0.5 is chance). In terms of discriminativity $d$, it was more difficult for the fine-tuned models to identify the correct feature when two or three features were identical across the pair (Table \ref{tab:pred-eval}). These average difficulties were driven by the failure on test sets where the non-contrastive features included shape (e.g., a pair showing a red~vs.~a blue block), indicating that the shape was easiest to pick up on for the models, although all features were mentioned in all training captions. For instance, $d$ was 0.750 for SP on the object color-scale matched test set, and 0.724 on the random two-feature test set, but 0.501 on the shape-object color matched set. The discriminativity on random and background feature matched three-feature test sets was 0.618 | 0.875 (RP) and 0.854 | 0.605 (SP), while it was only 0.087 (RP) and 0.164 (SP) on the object feature matched test set.

The better contrastive performance of the baseline came at a cost of generally overmodifying the messages with contrastive features (see low contrastive efficiency, Table~\ref{tab:pred-eval}). Low relevance scores also show that the baseline did not identify functionally appropriate features well. In contrast, both fine-tuned models showed higher contrastive efficiency and relevance, indicating that the task based fine-tuning might have helped the models to learn contrastiveness. 
The fine-tuned models also showed higher optimal constrastivity which is, however, still far from perfect. In general, no qualitative differences between the two- and three-feature datasets or RP and SP settings are apparent.  

Figure \ref{fig:byFt-biases} shows how frequently the models' predictions mentioned a specific feature when it was contrastively \textit{irrelevant} (i.e., it zooms in on predictions where $r < 1$). For the fine-tuned models, it suggests potential biases towards redundantly producing object-related features (shape, scale, color of object), matching human biases (see Section~\ref{section1}), as opposed to background descriptions. The proportions slightly increase for object color and scale in the two- and three-feature test sets, potentially hinting at overmodification as the model's loophole behavior in a more complex setting. The SP model has a stronger redundancy propensity than RP.  
The apparent trend towards mentioning shape is in line with the pattern of discriminativity results described above where models relied on the shape being the discriminative feature between target and distractor.

\section{Conclusion}
We provide the A3DS dataset alongside evaluation metrics for investigating referential pragmatic abilities acquired by grounded language models on this dataset. With this dataset, we identify that an image captioner fine-tuned interactively via reinforcement learning developed a strikingly human-like shape bias, while being less overinformative than a task-neutral model. Future research could expand such evaluations by including metrics which investigate additional aspects that might matter to human referential expression generation (e.g., the current metrics are agnostic to the surface order of discriminative features, while humans have preferences towards certain adjective ordering; \citet{scontras2017subjectivity}).  Although these results are specific to the given architecture, with this work we hope to inspire research opening up black box language models---an important task in the age of LLMs. 

\section*{Limitations}
The identified tendencies towards mentioning object-related features and the reliance on the shape as a contrastive feature might be driven by the grammatical structure of the annotations, mostly presenting object features in sentence-initial subject position, although 40\%  of exhaustive captions mention either the scale or the object color as the last word in the sentence. Therefore, these results call for investigating the biases of model architectures less sensitive to sentence length than LSTMs, as well as extending the annotations with additional grammars. Further, this evaluation provides descriptive results of the models' pragmatic abilities, leaving the question of whether it is indeed a pragmatic inductive bias or, e.g., structural language drift \citep{lazaridou-etal-2020-multi} causing the observed patterns, unanswered. Finally, since the evaluation pertains to the surface form of the predictions, applying decoding schemes other than greedy decoding used in this work might provide different patterns, indicating to which degree potential biases are due to model mechanics in opposition to sampling parameters.


\section*{Acknowledgements}
We would like to thank Elia Bruni for his support of the work which led to this paper, and Xenia Ohmer and Leon Schmid for helpful discussions. We also acknowledge support by the state of Baden-Württemberg through the computing resources provided by bwHPC and the German Research Foundation (DFG) through grant INST 35/1597-1 FUGG. Michael Franke is a member of the Machine Learning Cluster of Excellence, EXC number 2064/1 – Project number 39072764.
\bibliography{anthology,custom}

\begin{thebibliography}{25}
\expandafter\ifx\csname natexlab\endcsname\relax\def\natexlab#1{#1}\fi

\bibitem[{Andreas and Klein(2016)}]{andreas-klein-2016-reasoning}
Jacob Andreas and Dan Klein. 2016.
\newblock \href {https://doi.org/10.18653/v1/D16-1125} {Reasoning about
  pragmatics with neural listeners and speakers}.
\newblock In \emph{Proceedings of the 2016 Conference on Empirical Methods in
  Natural Language Processing}, pages 1173--1182, Austin, Texas. Association
  for Computational Linguistics.

\bibitem[{Banerjee and Lavie(2005)}]{banerjee-lavie-2005-meteor}
Satanjeev Banerjee and Alon Lavie. 2005.
\newblock \href {https://aclanthology.org/W05-0909} {{METEOR}: An automatic
  metric for {MT} evaluation with improved correlation with human judgments}.
\newblock In \emph{Proceedings of the {ACL} Workshop on Intrinsic and Extrinsic
  Evaluation Measures for Machine Translation and/or Summarization}, pages
  65--72, Ann Arbor, Michigan. Association for Computational Linguistics.

\bibitem[{Burgess and Kim(2018)}]{burgess20183d}
Chris Burgess and Hyunjik Kim. 2018.
\newblock 3d shapes dataset.
\newblock https://github.com/deepmind/3dshapes-dataset/.

\bibitem[{Cohn-Gordon et~al.(2018)Cohn-Gordon, Goodman, and
  Potts}]{cohn-gordon-etal-2018-pragmatically}
Reuben Cohn-Gordon, Noah Goodman, and Christopher Potts. 2018.
\newblock \href {https://doi.org/10.18653/v1/N18-2070} {Pragmatically
  informative image captioning with character-level inference}.
\newblock In \emph{Proceedings of the 2018 Conference of the North {A}merican
  Chapter of the Association for Computational Linguistics: Human Language
  Technologies, Volume 2 (Short Papers)}, pages 439--443, New Orleans,
  Louisiana. Association for Computational Linguistics.

\bibitem[{Degen et~al.(2020)Degen, Hawkins, Graf, Kreiss, and
  Goodman}]{degen2020redundancy}
Judith Degen, Robert~D Hawkins, Caroline Graf, Elisa Kreiss, and Noah~D
  Goodman. 2020.
\newblock \href {https://doi.org/10.1037/rev0000186} {When redundancy is
  useful: {A} {B}ayesian approach to “overinformative” referring
  expressions.}
\newblock \emph{Psychological Review}, 127(4):591.

\bibitem[{Goodman and Frank(2016)}]{goodman2016pragmatic}
Noah~D Goodman and Michael~C Frank. 2016.
\newblock \href {https://doi.org/10.1016/j.tics.2016.08.005} {Pragmatic
  language interpretation as probabilistic inference}.
\newblock \emph{Trends in cognitive sciences}, 20(11):818--829.

\bibitem[{Grice(1975)}]{grice1975logic}
Herbert~P Grice. 1975.
\newblock Logic and conversation.
\newblock In \emph{Speech acts}, pages 41--58. Brill.

\bibitem[{Havrylov and Titov(2017)}]{havrylov2017emergence}
Serhii Havrylov and Ivan Titov. 2017.
\newblock Emergence of language with multi-agent games: Learning to communicate
  with sequences of symbols.
\newblock \emph{Advances in neural information processing systems}, 30.

\bibitem[{Karpathy et~al.(2014)Karpathy, Joulin, and
  Fei-Fei}]{karpathy2014deep}
Andrej Karpathy, Armand Joulin, and Li~F Fei-Fei. 2014.
\newblock Deep fragment embeddings for bidirectional image sentence mapping.
\newblock \emph{Advances in neural information processing systems}, 27.

\bibitem[{Kim and Mnih(2018)}]{pmlr-v80-kim18b}
Hyunjik Kim and Andriy Mnih. 2018.
\newblock \href {https://proceedings.mlr.press/v80/kim18b.html} {Disentangling
  by factorising}.
\newblock In \emph{Proceedings of the 35th International Conference on Machine
  Learning}, volume~80 of \emph{Proceedings of Machine Learning Research},
  pages 2649--2658. PMLR.

\bibitem[{Kramer and van Deemter(2012)}]{KramerDeemter2012:Computational-G}
Emiel Kramer and Kees van Deemter. 2012.
\newblock Computational generation of referring expressions: {A} survey.
\newblock \emph{Computational Linguistics}, 38(1):173--218.

\bibitem[{Lake et~al.(2017)Lake, Ullman, Tenenbaum, and
  Gershman}]{lake_ullman_tenenbaum_gershman_2017}
Brenden~M. Lake, Tomer~D. Ullman, Joshua~B. Tenenbaum, and Samuel~J. Gershman.
  2017.
\newblock \href {https://doi.org/10.1017/S0140525X16001837} {Building machines
  that learn and think like people}.
\newblock \emph{Behavioral and Brain Sciences}, 40:e253.

\bibitem[{Lazaridou and Baroni(2020)}]{Lazaridou2020EmergentMC}
Angeliki Lazaridou and Marco Baroni. 2020.
\newblock Emergent multi-agent communication in the deep learning era.
\newblock \emph{ArXiv}, abs/2006.02419.

\bibitem[{Lazaridou et~al.(2016)Lazaridou, Peysakhovich, and
  Baroni}]{lazaridou2016multi}
Angeliki Lazaridou, Alexander Peysakhovich, and Marco Baroni. 2016.
\newblock Multi-agent cooperation and the emergence of (natural) language.
\newblock \emph{arXiv preprint arXiv:1612.07182}.

\bibitem[{Lazaridou et~al.(2020)Lazaridou, Potapenko, and
  Tieleman}]{lazaridou-etal-2020-multi}
Angeliki Lazaridou, Anna Potapenko, and Olivier Tieleman. 2020.
\newblock \href {https://doi.org/10.18653/v1/2020.acl-main.685} {Multi-agent
  communication meets natural language: Synergies between functional and
  structural language learning}.
\newblock In \emph{Proceedings of the 58th Annual Meeting of the Association
  for Computational Linguistics}, pages 7663--7674, Online. Association for
  Computational Linguistics.

\bibitem[{Lin(2004)}]{lin-2004-rouge}
Chin-Yew Lin. 2004.
\newblock \href {https://aclanthology.org/W04-1013} {{ROUGE}: A package for
  automatic evaluation of summaries}.
\newblock In \emph{Text Summarization Branches Out}, pages 74--81, Barcelona,
  Spain. Association for Computational Linguistics.

\bibitem[{Papineni et~al.(2002)Papineni, Roukos, Ward, and
  Zhu}]{papineni-etal-2002-bleu}
Kishore Papineni, Salim Roukos, Todd Ward, and Wei-Jing Zhu. 2002.
\newblock \href {https://doi.org/10.3115/1073083.1073135} {{B}leu: a method for
  automatic evaluation of machine translation}.
\newblock In \emph{Proceedings of the 40th Annual Meeting of the Association
  for Computational Linguistics}, pages 311--318, Philadelphia, Pennsylvania,
  USA. Association for Computational Linguistics.

\bibitem[{Scontras et~al.(2017)Scontras, Degen, and
  Goodman}]{scontras2017subjectivity}
Gregory Scontras, Judith Degen, and Noah~D Goodman. 2017.
\newblock Subjectivity predicts adjective ordering preferences.
\newblock \emph{Open Mind}, 1(1):53--66.

\bibitem[{Searle(1969)}]{searle1969speech}
John~R Searle. 1969.
\newblock \emph{Speech acts: An essay in the philosophy of language}, volume
  626.
\newblock Cambridge university press.

\bibitem[{Shen et~al.(2019)Shen, Fried, Andreas, and
  Klein}]{shen-etal-2019-pragmatically}
Sheng Shen, Daniel Fried, Jacob Andreas, and Dan Klein. 2019.
\newblock \href {https://doi.org/10.18653/v1/N19-1410} {Pragmatically
  informative text generation}.
\newblock In \emph{Proceedings of the 2019 Conference of the North {A}merican
  Chapter of the Association for Computational Linguistics: Human Language
  Technologies, Volume 1 (Long and Short Papers)}, pages 4060--4067,
  Minneapolis, Minnesota. Association for Computational Linguistics.

\bibitem[{Vedantam et~al.(2017)Vedantam, Bengio, Murphy, Parikh, and
  Chechik}]{vedantam2017context}
Ramakrishna Vedantam, Samy Bengio, Kevin Murphy, Devi Parikh, and Gal Chechik.
  2017.
\newblock Context-aware captions from context-agnostic supervision.
\newblock In \emph{Proceedings of the IEEE Conference on Computer Vision and
  Pattern Recognition}, pages 251--260.

\bibitem[{Vedantam et~al.(2015)Vedantam, Lawrence~Zitnick, and
  Parikh}]{vedantam2015cider}
Ramakrishna Vedantam, C~Lawrence~Zitnick, and Devi Parikh. 2015.
\newblock Cider: Consensus-based image description evaluation.
\newblock In \emph{Proceedings of the IEEE conference on computer vision and
  pattern recognition}, pages 4566--4575.

\bibitem[{Vinyals et~al.(2016)Vinyals, Toshev, Bengio, and
  Erhan}]{vinyals2016show}
Oriol Vinyals, Alexander Toshev, Samy Bengio, and Dumitru Erhan. 2016.
\newblock Show and tell: Lessons learned from the 2015 {MSCOCO} image
  captioning challenge.
\newblock \emph{IEEE transactions on pattern analysis and machine
  intelligence}, 39(4):652--663.

\bibitem[{Zarrie{\ss} et~al.(2021)Zarrie{\ss}, Buschmeier, Han, and
  Sch{\"u}z}]{zarriess-etal-2021-decoding}
Sina Zarrie{\ss}, Hendrik Buschmeier, Ting Han, and Simeon Sch{\"u}z. 2021.
\newblock \href {https://aclanthology.org/2021.inlg-1.41} {Decoding, fast and
  slow: A case study on balancing trade-offs in incremental, character-level
  pragmatic reasoning}.
\newblock In \emph{Proceedings of the 14th International Conference on Natural
  Language Generation}, pages 371--376, Aberdeen, Scotland, UK. Association for
  Computational Linguistics.

\bibitem[{Zhou et~al.(2020)Zhou, Palangi, Zhang, Hu, Corso, and
  Gao}]{zhou2020unified}
Luowei Zhou, Hamid Palangi, Lei Zhang, Houdong Hu, Jason Corso, and Jianfeng
  Gao. 2020.
\newblock Unified vision-language pre-training for image captioning and {VQA}.
\newblock In \emph{Proceedings of the AAAI Conference on Artificial
  Intelligence}, volume~34, pages 13041--13049.

\end{thebibliography}
\bibliographystyle{acl_natbib}




\end{document}